\documentclass[lettersize,journal]{IEEEtran}
\usepackage{amsmath,amsfonts}
\usepackage{multirow}
\usepackage{booktabs}
\usepackage[misc]{ifsym}
\usepackage{ntheorem}
\usepackage{float}
\usepackage{subfigure}
\usepackage{threeparttable}
\usepackage{algorithmic}
\usepackage{algorithm}
\usepackage{array}
\usepackage[caption=false,font=normalsize,labelfont=sf,textfont=sf]{subfig}
\usepackage{textcomp}
\usepackage{stfloats}
\usepackage{url}
\usepackage{verbatim}
\usepackage{graphicx}
\usepackage{cite}
\usepackage{caption}
\def\BibTeX{{\rm B\kern-.05em{\sc i\kern-.025em b}\kern-.08em
    T\kern-.1667em\lower.7ex\hbox{E}\kern-.125emX}}
\usepackage{balance}
\usepackage{colortbl}
\usepackage{diagbox}
\definecolor{mygray}{gray}{.7}

\newtheorem*{proof}{Proof}
\def\ie{{\em i.e.}}
\def\eg{{\em e.g.}}
\def\etal{{\em et al.}}

\graphicspath{{./figures/}}

\newcommand{\equref}[1]{(\ref{#1})}

\newcommand{\myPara}[1]{\noindent\textbf{#1}}

\newcommand{\mb}[1]{\mathbb{#1}}

\newtheorem{theorem}{Theorem}{}

{}

\begin{document}

\title{Learning Privacy-Preserving Student Networks via Discriminative-Generative Distillation}

\author{Shiming~Ge,~\IEEEmembership{Senior Member,~IEEE,}
        Bochao~Liu,
        Pengju~Wang,
        Yong~Li,
        and~Dan~Zeng~\IEEEmembership{Senior Member,~IEEE}
\thanks{Shiming Ge, Bochao Liu, Pengju~Wang and Yong Li are with the Institute of Information Engineering, Chinese Academy of Sciences, Beijing 100095, China, and with School of Cyber
Security at University of Chinese Academy of Sciences, Beijing 100049, China. Email: \{geshiming, liubochao, wangpengju, liyong\}@iie.ac.cn.}
\thanks{Dan~Zeng is with the School of Communication and Information Engineering, Shanghai University, Shanghai 200444, China. E-mail: dzeng@shu.edu.cn. Email: dzeng@shu.edu.cn.}
\thanks{Y. Li is the corresponding author. (e-mail: liyong@iie.ac.cn)} 
}

\markboth{IEEE Transactions on Image Processing}%
{Shell \MakeLowercase{\textit{et al.}}: A Sample Article Using IEEEtran.cls for IEEE Journals}


\maketitle

\begin{abstract}
While deep models have proved successful in learning rich knowledge from massive well-annotated data, they may pose a privacy leakage risk in practical deployment. It is necessary to find an effective trade-off between high utility and strong privacy. In this work, we propose a discriminative-generative distillation approach to learn privacy-preserving deep models. Our key idea is taking models as bridge to distill knowledge from private data and then transfer it to learn a student network via two streams. First, discriminative stream trains a baseline classifier on private data and an ensemble of teachers on multiple disjoint private subsets, respectively. Then, generative stream takes the classifier as a fixed discriminator and trains a generator in a data-free manner. After that, the generator is used to generate massive synthetic data which are further applied to train a variational autoencoder (VAE). Among these synthetic data, a few of them are fed into the teacher ensemble to query labels via differentially private aggregation, while most of them are embedded to the trained VAE for reconstructing synthetic data. Finally, a semi-supervised student learning is performed to simultaneously handle two tasks: knowledge transfer from the teachers with distillation on few privately labeled synthetic data, and knowledge enhancement with tangent-normal adversarial regularization on many triples of reconstructed synthetic data. In this way, our approach can control query cost over private data and mitigate accuracy degradation in a unified manner, leading to a privacy-preserving student model. Extensive experiments and analysis clearly show the effectiveness of the proposed approach.
\end{abstract}


\section{Introduction}
\IEEEPARstart{D}{eep} learning~\cite{lecun2015deep} has delivered impressive performance in image recognition ~\cite{Lecun98cnn,alexnet2012nips,vggnet2015iclr,resnet2016cvpr,dosovitskiy2020image} due to the powerful capacity of deep networks on learning rich knowledge from large-scale annotated data. However, the deployment of deep models may suffer from the leakage risk of data privacy. Recent works~\cite{fredrikson2015ccs,yang2019ccs} have shown that the private information in the training data can be easily recovered with a few access to the released model. Thus, many real-world requirements~\cite{abowd2018us,erlingsson2014rappor} need to provide high-performance models while protecting data privacy. Thus, it is necessary to explore a feasible solution that can address a key challenge for model deployment: how to effectively learn a  privacy-preserving deep model without remarkable loss of inference accuracy?  

\begin{figure}[!t]
\begin{center}
  \includegraphics[width=1.0\linewidth]{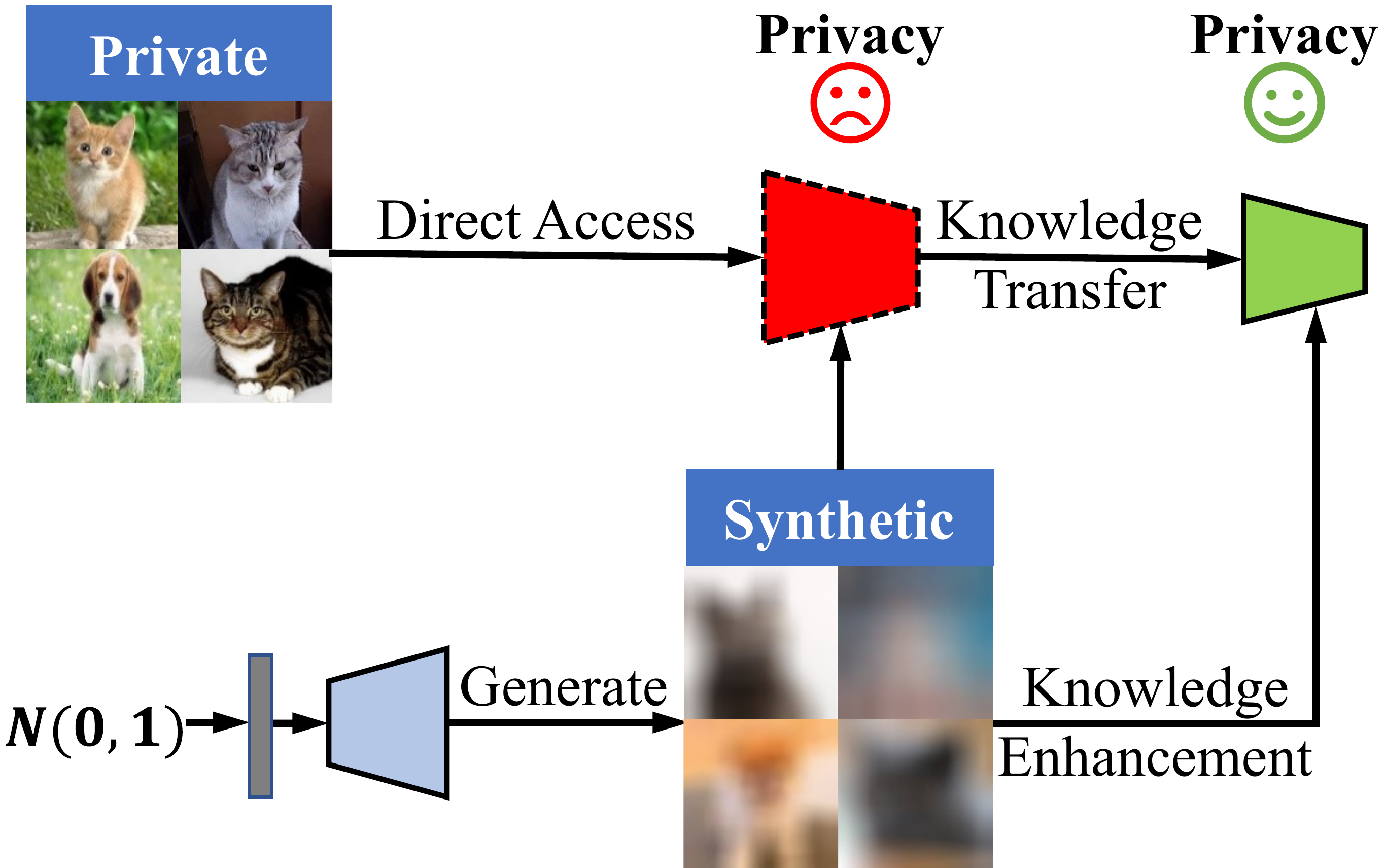}
  \caption{The models that are learned with direct access to private data may leak privacy. To address that, we aim to learn privacy-preserving models by knowledge transfer from directly-trained discriminative model(s), combined with knowledge enhancement using synthetic data generated by generative model.}
  \label{fig:motivation}
\end{center}
\end{figure}

Compared to traditional learning solutions that directly access to private data and lead to privacy leakage in released model (in red in Fig.~\ref{fig:motivation}), the privacy-preserving learning solutions usually add privacy protection strategies or avoid released model (in green in Fig.~\ref{fig:motivation}) directly access to private data during training. Towards this end, many existing approaches have been proposed, which are mainly based on differential privacy~\cite{dwork2006calibrating}. According to the privacy-preserving strategy, they can be roughly grouped into two categories: the implicit category and explicit category. 

The ``implicit'' approaches leverage differentially private learning to train models from private data by enforcing differential privacy on the weights or gradients during training. The prior approach is introducing differential privacy into stochastic gradient descent to train deep networks~\cite{abadi2016deep}. Later, Chen \etal~\cite{chen2020stochastic} proposed a stochastic variant of classic backtracking line search algorithm to reduce privacy loss. Papernot \etal~\cite{papernot2021tempered} proposed to replace the unbounded ReLU activation function with a bounded tempered sigmoid function to retain more gradient information. Some recent works proposed gradient operations for private learning by denoising~\cite{nasr2020improving}, clipping~\cite{chen2020nips}, perturbation~\cite{yu2021iclr} or compression~\cite{wang2021ccs}. Typically, these approaches have promising privacy protection, but often suffer from a big drop in inference accuracy over their non-private counterparts. By contrast, the ``explicit'' approaches pretrain models on private data and then use auxiliary public or synthetic data to learn the released models with knowledge transfer by enforcing differential privacy on the outputs of pretrained models~\cite{papernot2016semi,hamm2016learning,papernot2018scalable,zhu2020Private}. Papernot \etal~\cite{papernot2016semi} proposed private aggregation of teacher ensembles (PATE) to learn a student model by privately transferring the teacher knowledge with public data, while Hamm \etal~\cite{hamm2016learning} proposed a new risk weighted by class probabilities estimated from the ensemble to reduce the sensitivity of majority voting. Zhu \etal~\cite{zhu2020Private} proposed a practically data-efficient scheme based on private release of k-nearest neighbor queries, which can avoid the decline of accuracy caused by partitioning training set. These approaches generally can improve the model performance while massive unlabeled public data with the same distribution as private data are available. However, these public data are difficult to obtain and the models trained in this way are still at risk of malicious attacks. Recent works~\cite{xie2018differentially,jordon2018pate} trained a private generator and used the generated synthetic data to replace the auxiliary public data. Generally speaking, the most important process in the explicit category is transferring sufficient knowledge from private data to auxiliary data with minimal privacy leakage. Thus, the key issue that needs to be carefully addressed is applying \emph{reliable models} to extract knowledge from private data and exploring \emph{effective auxiliary data} to transfer knowledge.  

\begin{figure*}[!ht]
	\begin{center}
		\includegraphics[width=1\linewidth]{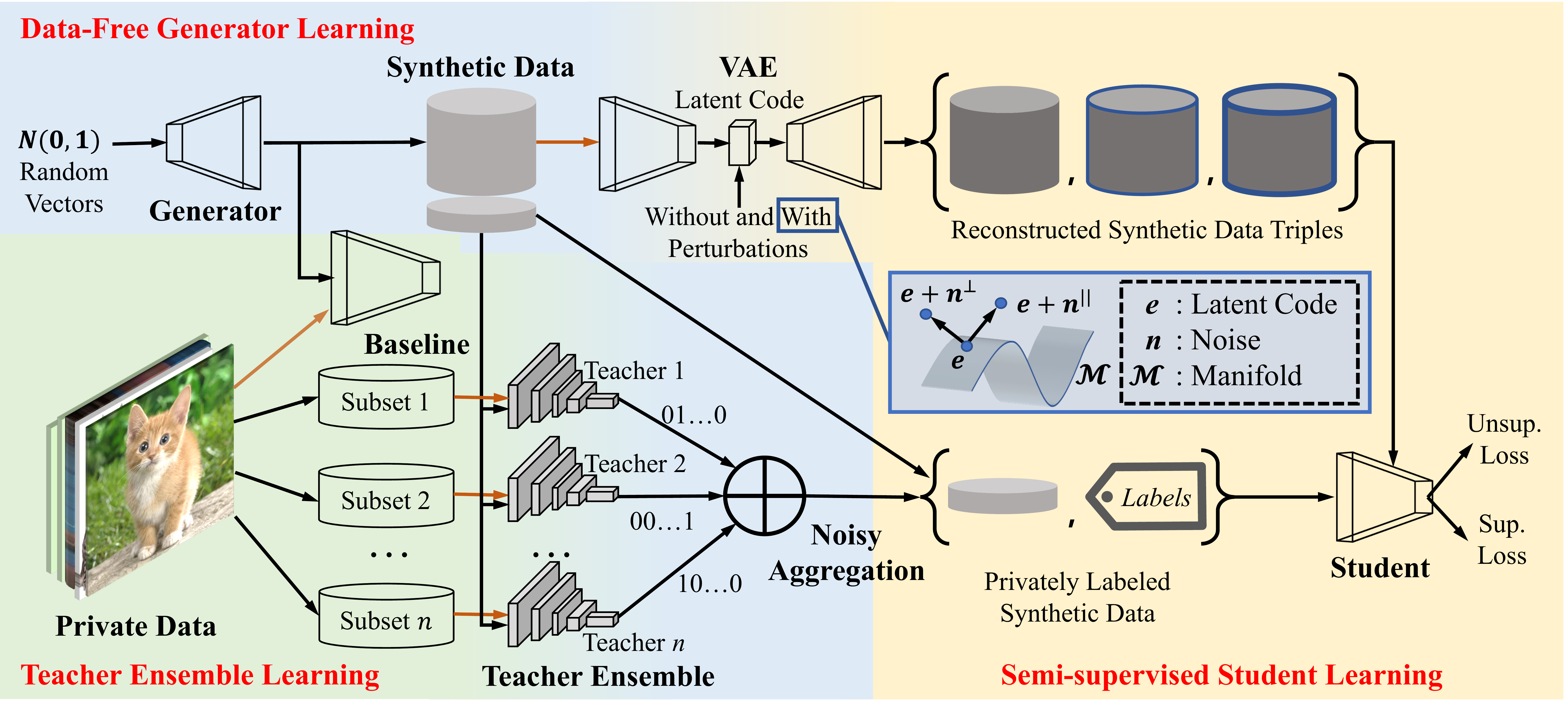}
	\end{center}
	\caption{Overview of the approach. The privacy-preserving student learning is performed with discriminative-generative distillation via two streams. First, discriminative stream trains a baseline classifier on private data and an ensemble of multiple teachers on disjoint private subsets, and generative stream takes the baseline as a fixed discriminator to train a generator in a data-free manner. The generator is employed to generate massive synthetic data that are used to pretrain a variational autoencoder (VAE). Then, the synthetic data are splitted into two parts: a few of them are fed into teacher ensemble in discriminative stream to query labels by noisy aggregation, and most of them are embedded into the VAE space and reconstructed with and without latent code perturbation. Finally, two streams converge to perform semi-supervised student learning by transferring teacher knowledge with few labeled synthetic data and regularizing with massive VAE-reconstructed synthetic data.}	
	\label{fig:framework}
\end{figure*}

Inspired by this fact, we propose a teacher-student learning approach to train privacy-preserving student networks via discriminative-generative distillation, which applies \emph{discriminative and generative models} to distill private knowledge and then explores \emph{generated synthetic data} to perform knowledge transfer (Fig.~\ref{fig:motivation}). The objective is to enable an effective learning that achieves a promising trade-off between high model utility and strong privacy protection. As shown in Fig.~\ref{fig:framework}, the student is trained by using two streams. First, discriminative stream trains a baseline classifier on all private data and an ensemble of separate teachers on disjoint subsets, while generative stream takes the baseline classifier as a fixed discriminator and trains a generator in a data-free manner. Massive synthetic data are then generated with the generator and used to train a variational autoencoder (VAE)~\cite{kingma2013auto}. After that, a few of the synthetic data are fed into the teacher ensemble to query labels with Laplacian aggregation, while most of the synthetic data are fed into VAE to achieve massive data triples by perturbing the latent codes. Finally, a semi-supervised learning is performed by simultaneously handling two tasks: knowledge transfer via supervised data classification, and knowledge enhancement via self-supervised model regularization. 

In summary, our approach can effectively learn privacy-preserving student networks by three key components. First, data-free generator learning is incorporated to generate massive synthetic data. These synthetic data are difficult to be identified from appearance but have similar distribution with private data in discriminative space. Therefore, the student learning does not involve any private data and the synthetic data do not expose the information of private data even if they are recovered. Second, differential privacy is incorporated to provide a strong privacy guarantee theoretically. In Laplacian aggregation of teacher ensemble, student’s access to its teachers is limited by reducing label queries, so that the student’s exposure to teachers’ knowledge can be meaningfully quantified and bounded. Third, tangent-normal adversarial regularization is adopted to improve the capacity and robustness of student. In semi-supervised student learning, synthetic data are embedded into the pretrained VAE space and reconstructed from latent codes by adding perturbation along both tangent and normal directions of distribution manifold. Then, the tangent regularization can enforce the local smoothness of the student along the underlying manifold and improve model accuracy, while the normal regularization imposes robustness on the student against noise. In this way, the two regularization terms complement each other, jointly facilitating knowledge transfer from the teacher ensemble to the student. Our approach provides a unified framework to learn privacy-preserving student networks. The data-free generator learning and Laplacian aggregation can protect the private data, and adversarial regularization via VAE reconstruction of the synthetic data can better learn data manifold. Combining them together can protect private data while reducing the impact of noisy labels and instability of the synthetic data.

Our major contributions are three folds: 1) we propose a discriminative-generative distillation approach to train privacy-preserving student networks that achieves an effective trade-off between high utility and strong privacy, 2) we propose to combine data-free generator learning and VAE-based model regularization which facilitates knowledge transfer in a semi-supervised manner, and 3) we conduct extensive experiments and analysis to demonstrate the effectiveness of our approach.

\section{Related Works}
The approach we proposed in this paper aims to learn privacy-preserving student networks by distilling knowledge from private data and transferring it to synthetic data. Therefore, we briefly review related works from three aspects, including differentially private learning, learning with synthetic data and teacher-student learning.

\subsection{Differentially Private Learning} 
Differentially private learning~\cite{liu2022acs} aims to address tasks like healthcare~\cite{suriyakumar2021chasing} where the data are private and the learning process meets differential privacy requirements. Differential privacy provides a guarantee that two adjacent databases produce statistically indistinguishable results under a reasonable privacy budget. 

Previous works~\cite{pathak2011privacy,bassily2014private} considered using differential privacy in machine learning settings. Shokri \etal~\cite{shokri2015privacy} introduced a privacy-preserving distributed stochastic gradient descent (SGD) algorithm which applies to non-convex models. Its privacy bound is decided by the number of model parameters that are related to the representation ability of the model, leading to an inefficient trade-off between privacy and model capacity. Abadi \etal~\cite{abadi2016deep} provided a stricter bound on the privacy loss induced by a noisy SGD by introducing moments accountant. Papernot \etal~\cite{papernot2016semi} proposed a general framework named private aggregation of teacher ensembles (PATE) for private training. PATE uses semi-supervised learning to transfer the knowledge of the teacher ensemble to the student by using a differentially private aggregation. It uses the assumption that the student has access to additional unlabeled data. To reduce erroneous aggregation results, Xiang \etal~\cite{xiang2020achieving} proposed a private consensus protocol by returning only the highest voting results above a threshold in aggregation of teacher ensembles, leading to accuracy improvement under the same privacy level. Gao \etal~\cite{gao2022sedml} improved PATE to securely and efficiently harness distributed knowledge by using lightweight cryptography, which can achieve strong protection for individual labels. Miyato \etal~\cite{miyato2018virtual} proposed virtual  adversarial training to avoid the requirements of label information, which reduces queries to the privacy model and protects data privacy. Jagannathan \etal~\cite{jagannathan2013semi} combined Laplacian mechanism with decision trees and proposed a random forest algorithm to protect privacy. The idea of differentially private learning can suggest the usage of data for training models under a certain privacy budget. 

\subsection{Learning with Synthetic Data}
With the development of generative adversarial networks (GANs)~\cite{goodfellow2014nips}, recent works began to use synthetic data in training deep networks. Zhang \etal~\cite{dai2017good} found that the performance of classifiers trained in a semi-supervised manner using synthetic data could not be guaranteed and proposed Bad GAN to preferentially select the generator, which greatly improves the feature matching of GANs. Dumoulin \etal~\cite{dumoulin2016adversarially} proposed to jointly learn a generation network and an inference network using synthetic data generated by generation network, achieving a very competitive performance. Salimans \etal~\cite{salimans2016improved} presented a variety of architectural features and training procedures, which improves the performance of both classifier and generator. Kumar \etal~\cite{kumar2017semi} proposed to estimate the tangent space to the data manifold using GANs and employ it to inject invariances into the classifier, which can greatly improve in terms of semantic similarity of the reconstructed samples with the input samples. Luo \etal~\cite{luo2018smooth} introduced smooth neighbors on teacher graphs, which improves the performance of classifier through the implicit self-ensemble of models. Qi \etal~\cite{qi2018global} presented localized GAN to learn the manifold of real data, which could not only produce diverse image transformations but also deliver superior classification performance. The works~\cite{beaulieu2019privacy,torkzadehmahani2019dp,xie2018differentially,zhang2018differentially} used differentially private stochastic gradient descent~(DPSGD) to train GANs, which has been proven effective in generating high-dimensional sanitized data~\cite{torkzadehmahani2019dp}. However, DPSGD relies on carefully tuning of the clipping bound of gradient norm, \ie, the sensitivity value. Specifically, the optimal clipping bound varies greatly with model architecture and training dynamics, making the implementation of DPSGD difficult. In order to solve this problem, Chen \etal~\cite{chen2020gs} used Wasserstein GANs~\cite{arjovsky2017wasserstein,gulrajani2017improved} for a precise estimation of the sensitivity value, avoiding the intensive search of hyper-parameters while reducing the clipping bias. Generally, these approaches aim to generate synthetic data to facilitate model learning, while the privacy issue introduced by generated data is less considered.   

\subsection{Teacher-Student Learning}
Typically, teacher-student learning applies knowledge distillation~\cite{bucila2006model,hinton2015distilling,gou2021ijcv} to learn a more compact student model by mimicking the behaviors of a complex teacher model. It is used for model compression while hardly degrading the model performance. In the vanilla knowledge distillation, by using the softmax output of the teacher network as soft labels instead of hard class labels, the student model can learn how the teacher network behaves given tasks in a compact form. Since then, many works~\cite{romero2014fitnets,liu2019knowledge,park2019relational,xue2019transferable,shen2019meal} had used and improved this training method. Romero \etal~\cite{romero2014fitnets} proposed to add an additional linear projection layer. Tian \etal~\cite{tian2020iclr} proposed to combine contrastive learning with knowledge distillation. The teacher-student learning manner has been applied in many applications, such as low-resolution face recognition~\cite{ge2020bd}, action recognition~\cite{liu2021tip}, semantic segmentation~\cite{feng2021tip}, data generation~\cite{lin2020tip} and molecular generation~\cite{wang2021nature}. For circumstances when training data for the teacher are unavailable in practical problems such as privacy, Chen \etal~\cite{chen2019data} proposed a data-free knowledge distillation framework. It regards the pretrained teacher networks as a fixed discriminator and trains a generator to synthesize training samples for the student. To protect the privacy of the data, some works utilize structural improvements~\cite{hamm2016learning}, such as training a collection of teacher models~\cite{dietterich2000ensemble}. Recently, the distillation idea is used to control privacy loss~\cite{papernot2016semi,jordon2018pate}. The key issue in learning privacy-preserving models with distillation is to make knowledge transfer adequately and privately.

\section{The Approach}

\subsection{Problem Formulation}
Given a private dataset $\mathcal{D}$, the objective is learning a privacy-preserving student $\phi_s$ that does not reveal data privacy and has the capacity approximating to the baseline model $\phi_b$ trained directly on $\mathcal{D}$. To achieve that, we introduce both discriminative and generative models to enforce the knowledge transfer via  discriminative-generative distillation with two streams. Discriminative stream partitions $\mathcal{D}$ into $n$ disjoint subsets $\mathcal{D} = \left\{\mathcal{D}_{i}\right\}_{i=1}^{n}$ and learns an ensemble of multiple teachers $\phi_t=\{\phi_{t,i}\}_{i=1}^{n}$ where $\phi_{t,i}$ is trained on $\mathcal{D}_{i}$. Generative stream takes $\phi_b$ as a fixed discriminator and learns a generator $\phi_g$ to generate massive synthetic data $\hat{\mathcal{D}}$. A VAE $\{\phi_{e},\phi_{d}\}$ is pretrained on synthetic data, where $\phi_{e}$ and $\phi_{d}$ are the encoder and decoder respectively. The pretrained VAE is used to obtain data distribution information to facilitate model learning. To reduce the privacy budget, only a few of synthetic data $\hat{\mathcal{D}}_s\subset\hat{\mathcal{D}}$ are used to query the teacher ensemble and get the noisy labels $\hat{\mathcal{L}}_s$. The other unlabelled data $\hat{\mathcal{D}_u} = \hat{\mathcal{D}} \setminus \hat{\mathcal{D}_s}$ with $|\hat{\mathcal{D}}_u| \gg |\hat{\mathcal{D}}_s|$ are employed to provide manifold regularization, with the help of VAE. Thus, the student learning can be formulated by minimizing an energy function $\mathbb{E}$:
\begin{equation}
\begin{aligned}
\mathbb{E}(\mathbb{W}_s;\hat{\mathcal{D}}) &= 
\mathbb{E}_s(\phi_s(\mathbb{W}_s;\hat{\mathcal{D}}_s);\hat{\mathcal{L}}_s) + \\ 
&\mathbb{E}_u(\phi_s(\mathbb{W}_s;\phi_d(\phi_e(\hat{\mathcal{D}}_u))), \phi_s(\mathbb{W}_s;\phi_d(\mathbb{P}[\phi_e(\hat{\mathcal{D}}_u))])),
\end{aligned}
\label{eq:problem}
\end{equation}
where $\mathbb{W}_s$ is the parameters of student, $\mathbb{P}[\cdot]$ is the perturbation operator, $\mathbb{E}_s$ and $\mathbb{E}_u$ are supervised energy term and unsupervised energy term, respectively. 

We can see that the risk of privacy leakage can be effectively suppressed due to the isolation between the released student model and private data. The supervised energy term can enforce knowledge transfer on the class-related characteristics, while the unsupervised energy term performs self-supervised regularization to enhance knowledge. Towards this end, we solve Eq.~\equref{eq:problem} via three steps, including: 1) data-free generator learning to get synthetic data $\hat{\mathcal{D}}$ and train a VAE $\{\phi_{e},\phi_{d}\}$, 2) teacher ensemble learning to achieve the labels $\hat{\mathcal{L}}_s$ by differentially private aggregation, and 3) semi-supervised student learning to get $\mathbb{W}_s$.

\subsection{Data-Free Generator Learning}
Knowledge transfer from the model trained on private data or synthetic data generated by GAN pretrained on private data may lead to privacy leakage. Meanwhile, the models trained on public data may cause significant accuracy degradation due to distribution mismatch, since finding public data that match the distribution with private data often is very difficult~\cite{papernot2016semi}. Moreover, the resulting models are vulnerable to attacks since adversaries can also access public data. Thus, we aim to learn a generator in a data-free manner 
to generate synthetic data, which does not compromise privacy to assist in knowledge transfer from private data to learn released models. 

Unlike traditional GAN training where the discriminator is an online learned two-class classifier, our data-free generator learning first pretrains a baseline multi-class classifier $\phi_b$ (with parameters $\mathbb{W}_b$) on private data $\mathcal{D}$ that serves as the fixed discriminator, and then train the generator $\phi_g$ without data. It is suggested that the tasks of discrimination and classification can improve each other and the multi-class classifier can learn the data distribution better than the two-class discriminator \cite{odena2016semi,chen2019data}. Thus, the key of using the multi-class classifier as discriminator is defining a loss to evaluate the generated data. Towards this end, we assess a synthetic example $\mathbf{x}=\phi_g(\mathbb{W}_g;\mathbf{z})$ generated by $\phi_g$ with parameters $\mathbb{W}_g$ from a random vector $\mathbf{z}$ by the following loss:
\begin{equation}\label{eq:generatorloss}
\begin{aligned}
\mathcal{L}(\mathbf{x})=&\ell(\phi_b(\mathbb{W}_b;\mathbf{x}),\arg \max\limits_j(\phi_b(\mathbb{W}_b;\mathbf{x}))_j) + \\
& \alpha\phi_b(\mathbb{W}_b;\mathbf{x})\log{\phi_b(\mathbb{W}_b;\mathbf{x})} + \beta\|\phi_b(\mathbb{W}_b^{-};\mathbf{x})\|_{1},
\end{aligned}
\end{equation}
where $\alpha$ and $\beta$ are the tuning parameters to balance the effect of three terms, and we set them as 5 and 0.1 respectively. The first term $\ell(.)$ is cross entropy function that measures the one-hot classification loss, which enforces the generated data having similar distribution as the private data. The second term is the information entropy loss to measure the class balance of generated data. The third term uses $l_1$-norm $\|*\|_1$ to measure the activation loss, since the features $\phi_b(\mathbb{W}_b^{-};\mathbf{x})$ that are extracted by the discriminator and correspond to the output before the fully-connected layer tend to receive higher activation value if input data are real rather than some random vectors, where $\mathbb{W}_b^{-}\subset\mathbb{W}_b$ is the discriminator's backbone parameters. Then, using the fixed discriminator, the generator learning is performed iteratively via five steps: 
\begin{itemize}
    \item randomly generate a batch of noise vectors :~$\left\{\mathbf{z}_i\right\}^{m}_{i=1}$.
    \item generate synthetic samples $\left\{\mathbf{x}_i\right\}^{m}_{i=1}$ for training:~$\mathbf{x}_i=\phi_g(\mathbb{W}_g;\mathbf{z}_i)$.
    \item apply the discriminator on the mini-batch: $\mathbf{y}_i=\phi_b(\mathbb{W}_b;\mathbf{x}_i)$.
    \item calculate the loss function with Eq.~\equref{eq:generatorloss} on mini-batch: $\sum_{i}{\mathcal{L}(\mathbf{x}_i)}$.
    \item update weights $\mathbb{W}_g$ using back-propagation.
\end{itemize}

In this way, the synthetic data $\hat{\mathcal{D}}$ generated by the learned generator have a similar distribution to private data without compromising privacy. Fig.~\ref{fig:generator} shows some examples. The synthetic data are very helpful for student learning, which can greatly improve accuracy compared to using public data and reduce accuracy loss compared to using private data directly. With $\hat{\mathcal{D}}$, we train a VAE $\{\phi_e, \phi_d\}$, where the encoder $\phi_e$ with parameters $\mathbb{W}_e$ and decoder $\phi_d$ with parameters $\mathbb{W}_d$ are constructed with convolutional neural networks like~\cite{yu2019tangent}.

\subsection{Teacher Ensemble Learning}
Instead of using a single model as teacher that may lead to privacy leakage~\cite{dietterich2000ensemble}, we learn an ensemble of teachers for knowledge transfer. Towards this end, we partition the private data $\mathcal{D}$ into $n$ disjoint subsets $\left\{\mathcal{D}_{i}\right\}_{i=1}^{n}$ and then separately train a teacher $\phi_{t,i}$ with parameters $\mathbb{W}_{t,i}$ on each subset $\mathcal{D}_{i}$, leading to the teacher ensemble $\phi_t=\{\phi_{t,i}\}_{i=1}^n$.

In general, the number of teachers $n$ has an impact on knowledge extraction from private data. When $n$ is too large, the amount of each training subset data gets less and the teachers may be underfitted. When $n$ is too small, it will make the noise of differential privacy more influential and lead to unusable aggregated labels. Thus, the teacher number $n$ should be carefully set in experience.   

The teacher ensemble serves to label query, where the synthetic data $\mathbf{x} \in \hat{\mathcal{D}}_s$ is fed and the predicted labels by multiple teachers are privately aggregated: 
\begin{equation}\label{eq:Laplacian}
l=\arg\max\limits_k\left\{\mathcal{V}_{k}(\{\phi_{t,i}(\mathbb{W}_{t,i};\mathbf{x})\}_{i=1}^{n})+ Lap(2/\varepsilon_0)\right\},
\end{equation}
where $\mathcal{V}_k(\cdot)$ counts the votes of the query being predicted as class $k$ by all $n$ teachers, the final predicted label $l$ is noisy and used to supervise the student training, a low privacy budget $\varepsilon_0$ is used to adjust privacy protection and $Lap(2/\varepsilon_0)$ denotes the Laplacian distribution with location $0$ and scale $2/\varepsilon_0$.
For student training, each example from the query data $\hat{\mathcal{D}}_{s}$ is fed into the teacher ensemble and then the prediction is privately aggregated via Laplacian aggregation in Eq.~\equref{eq:Laplacian}, leading to $\hat{\mathcal{L}}_s=\{l_i\}_{i=1}^{|\hat{\mathcal{D}}_s|}$. Directly using the maximum value of vote counts as labels may leak privacy, so we add random noise to the voting results to introduce ambiguity. Intuitively, this means that multiple teachers jointly determine the query result, making it difficult for adversary to recover the training data. In addition to this, our approach can provide the same or stronger privacy guarantee than many state-of-the-arts~\cite{papernot2016semi,xie2018differentially,jordon2018pate,long2019scalable,harder2020differentially,chen2020gs,yu2021iclr,wang2021ccs} while reducing accuracy degradation by knowledge enhancement with an extra model regularization. It also means that our approach will have a less privacy cost when delivering student models with the same accuracy.

\subsection{Semi-Supervised Student Learning}
To reduce privacy leakage, we only use a few of synthetic data $\hat{\mathcal{D}}_s$ for label query. Thus, the teacher knowledge that transfers from private data to $\hat{\mathcal{D}}_s$ is not only noisy due to Laplacian aggregation but also insufficient due to limited data. To enhance knowledge transfer, we learn the student in a semi-supervised fashion by adding another unsupervised pathway. 
Each synthetic example $\mathbf{x}_j\in\hat{\mathcal{D}}_u$ is embedded into VAE space and get a mean vector $\mathbf{\mu}_j$ and a standard deviation vector $\mathbf{\sigma}_j$ with $\{\mathbf{\mu}_j,\mathbf{\sigma}_j\}=\phi_e(\mathbb{W}_e;\mathbf{x}_j)$ that form a normal distribution $\mathcal{N}(\mathbf{\mu}_j,\mathbf{\sigma}_j)$. Then, the data is reconstructed from a sampled code $\mathbf{e}_j$ as well as its perturbed versions along tangent and normal directions of the distribution manifold, leading to massive data triples $\mathcal{T}=\{(\hat{\mathbf{x}}_j,\hat{\mathbf{x}}_{j}^{\parallel},\hat{\mathbf{x}}_{j}^{\perp})\}_{j=1}^{|\hat{\mathcal{D}}_u|}$ with:
\begin{equation}\label{VAEPair}
\hat{\mathbf{x}}_j=\phi_d(\mathbb{W}_d;\mathcal{M}(\mathbf{e}_j)),~\hat{\mathbf{x}}_{j}^{*}=\phi_d(\mathbb{W}_d;\mathcal{M}(\mathbf{e}_j+\mathbf{n}_{j}^{*})),
\end{equation}
where the mapping operator $\mathcal{M}(\cdot)$ projects the code into decoder input, $\mathbf{n}_{j}^{*}$ is random perturbation noise along tangent direction ($*=\parallel$) or normal direction ($*=\perp$). Then, the semi-supervised student learning is performed with $\{\hat{\mathcal{D}}_s,\hat{\mathcal{L}}_s\}$ and  $\mathcal{T}$. The supervised energy in Eq.~\equref{eq:problem} can be formulated as
\begin{equation}\label{eq:sup}
\mathbb{E}_s = \sum_{i=1}^{|\hat{\mathcal{D}}_s|}{\ell(\phi_s(\mb{W}_s;\mathbf{x}_i),l_i)},~s.t.~ \mathbf{x}_i \in \hat{\mathcal{D}}_s,~l_i \in \hat{\mathcal{L}}_s,
\end{equation}
and the unsupervised energy is formulated as 
\begin{equation}\label{eq:unsup}
\begin{aligned}
\mathbb{E}_u &= \sum_{j=1}^{|\hat{\mathcal{D}}_u|}{\|\phi_s(\mb{W}_{s}^{-};\hat{\mathbf{x}}_j)-\phi_s(\mb{W}_{s}^{-};\hat{\mathbf{x}}_{j}^{\perp})\|^2} \\
& +\sum_{j=1}^{|\hat{\mathcal{D}}_u|}{\|\phi_s(\mb{W}_{s}^{-};\hat{\mathbf{x}}_j)-\phi_s(\mb{W}_{s}^{-};\hat{\mathbf{x}}_{j}^{\parallel})\|^2} \\
& +\sum_{j=1}^{|\hat{\mathcal{D}}_u|}{\phi_s(\mb{W}_{s};\hat{\mathbf{x}}_j)\log{\phi_s(\mb{W}_{s};\hat{\mathbf{x}}_j)}},
\end{aligned}
\end{equation}
where $\mb{W}_{s}^{-}\subset\mb{W}_{s}$ is backbone parameters of the student for extracting features. We can see that the unsupervised energy Eq.~\equref{eq:unsup} includes normal regularization, tangent regularization and entropy regularization. The first two regularization terms enhance model robustness against perturbations along orthogonal and parallel directions to the underlying data manifold respectively, while entropy regularization ensures the student output more determinate predictions. This tangent-normal adversarial regularization by adding perturbation to the latent layer can make the student vary smoothly along tangent space and have strong robustness along normal space~\cite{yu2019tangent}.

The total energy comes from two streams. Discriminative stream employs the supervised energy for knowledge transfer from teacher ensemble with a few of query examples, while generative stream takes the unsupervised energy for knowledge enhancement via model regularization. The differentially private aggregation provides privacy protection, while the usage of VAE embedding and reconstruction can obtain the characteristics of the data in the tangent and normal spaces. In particular, generative stream applies self-supervised learning from massive unannotated data to compensate the knowledge that may miss in discriminative stream. 
\begin{table*}[!htbp]
 	\setlength{\tabcolsep}{10.8pt}%
 	\renewcommand\arraystretch{1.3}
	\begin{center}
		\begin{threeparttable}
		\caption{Performance comparisons with 6 explicit approaches: Test accuracy (\%) and the drop ($\nabla$) with respect to baseline under different privacy budget $\varepsilon$ ($\delta=10^{-5}$, BL: Baseline).}
			\begin{tabular}{@{}c|ccc|ccc@{}}
				\hline
				Approach&
				$\varepsilon$ &MNIST~(BL:~99.2~$\nabla$) &FMNIST~(BL:~91.0~$\nabla$)&$\varepsilon$&MNIST~(BL:~99.2~$\nabla$)&FMNIST~(BL:~91.0~$\nabla$)\cr
				\hline
				\multirow{1}{*}{DP-GAN~\cite{xie2018differentially}}&10.0&60.1 (39.1)&50.9 (40.1)&1.00&40.3 (58.9)&10.5 (80.5)\cr
			   \multirow{1}{*}{PATE-GAN~\cite{jordon2018pate}}&10.0&66.7 (32.5)&62.2 (28.8)&1.00&41.7 (57.5)&42.2 (48.8)\cr
				\multirow{1}{*}{G-PATE~\cite{long2019scalable}}&10.0&80.9 (18.3)&69.3 (21.7)&1.00&58.8 (40.4)&58.1 (32.9)\cr
				\multirow{1}{*}{DP-MERF~\cite{harder2020differentially}}&10.0&68.7 (30.5)&62.5 (28.5)&1.00&65.0 (34.2)&61.0 (30.0)\cr
				\multirow{1}{*}{GS-WGAN~\cite{chen2020gs}}&10.0&80.0 (19.2)&65.0 (26.0)&1.00&14.3 (84.9)&16.6 (74.4)\cr
		   	\multirow{1}{*}{DataLens~\cite{wang2021ccs}}&10.0&80.7 (18.5)&70.6 (20.4)&1.00&71.2 (28.0) &{64.8} ({26.2})\cr
				\multirow{1}{*}{\textbf{Our DGD}}&10.0&{\bf 97.4} ({\bf1.80})&{\bf 88.2} ({\bf 2.80})&1.00&{\bf 73.6} ({\bf 25.6})& {\bf 64.9} ({\bf 26.1})\cr
				\hline
			\end{tabular}
			\label{tab:mnist_and_fmnist}
		\end{threeparttable}
	\end{center}
\end{table*}

\subsection{Discussion\label{sec:discussion}}
\myPara{Practical Deployment.}~To learn a privacy-preserving student, our approach trains it from synthetic data generated with a generator pretrained in a data-free manner. Typically, the learning could be deployed in a single server where the private data are partitioned into several subsets to train an ensemble of teachers. Moreover, the learning is also suitable to deploy for jointly training models from distributed clients via a trusted server as the coordinator. In this case, each client trains a teacher on its private local data and all teachers form the teacher ensemble, while the trusted server aggregates the local data via centered learning or local knowledge via federated learning~\cite{FL2019} to pretrain a baseline classifier that is used to train a generator in a data-free manner. Then, the server applies the generator to generate massive synthetic data that are used to pretrain a VAE. After that, the server splits the synthetic data into two parts: a few of them are distributed to local clients to query labels by noisy aggregation in discriminative stream, and most of them are fed into generative stream for VAE reconstruction to get massive synthetic data triples. Finally, the student is trained on noisy labels and synthetic data triples in a semi-supervised manner within the trusted server. By allowing only student to be accessible to adversaries, the trained student could be deployed on practical applications and gives the differential privacy guarantee, introduced next.

\myPara{Privacy Analysis.}~According to the learning process in two streams, the total privacy budget contains two parts. Discriminative privacy budget is computed as PATE~\cite{papernot2016semi, papernot2018scalable}, achieving $\varepsilon_0$-differential privacy via Eq.~\equref{eq:Laplacian} and getting $(|\hat{\mathcal{D}}_s|\varepsilon_0^2+\varepsilon_0\sqrt{-2|\hat{\mathcal{D}}_s|\log{\delta}},\delta)-$differential privacy over $|\hat{\mathcal{D}}_s|$ queries for all $\delta \in (0,1)$~\cite{dwork2014algorithmic}. Generative privacy budget is computed according to the latent code perturbation in VAE construction. By taking the synthetic data in generative stream as a sequence, we achieve $\varepsilon_1$-differential privacy by adding Laplacian noise with scale $2c/\varepsilon_1$ to the normalized latent codes, where $c$ is the dimension of latent codes. It could be explained as follow. According to Laplacian mechanism and post-processing theorem~\cite{dwork2014algorithmic}, we have: for any two different images $\mathbf{x}_j$ and $\mathbf{x}_j^{\prime}$ as well as possible reconstructed output $\hat{\mathbf{x}}_j$, the VAE reconstruction mechanism $\mathcal{A}$ satisfies  $\operatorname{Pr}[\mathcal{A}(\mathbf{x}_j)=\hat{\mathbf{x}}_j] \leq \exp(\varepsilon_1) \cdot \operatorname{Pr}\left[\mathcal{A}\left(\mathbf{x}_j^{\prime}\right)=\hat{\mathbf{x}}_j\right]$ where $\operatorname{Pr}[\cdot]$ is the probability function. Then, we have the theorem.
\begin{theorem}\label{theorem:VAE}
\itshape{The sequence of VAE reconstruction mechanism $\mathcal{A}$, denoted as $\mathcal{A}(\hat{\mathcal{D}}_u)$ satisfies $\varepsilon_1$-differential privacy.}
\begin{proof}
\itshape{For any two adjacent datasets $\hat{\mathcal{D}}_u$ and $\hat{\mathcal{D}}_u^{\prime}$ where ${\mathbf{x}_j}\in\hat{\mathcal{D}}_u$ and ${\mathbf{x}_j^{\prime}}\in\hat{\mathcal{D}}_u^{\prime}$ are the two only different images, we have
\begin{equation}\label{eq:theorem}
\begin{aligned}
\operatorname{Pr}&\left[ \mathcal{A}\left(\hat{\mathcal{D}}_u\right) \subseteq \mathcal{O}\right] \\
=& \operatorname{Pr}\left[\mathcal{A}\left(\hat{\mathcal{D}}_u \cap \hat{\mathcal{D}}_u^{\prime}\right)\subseteq \mathcal{O}\right] \cdot \operatorname{Pr}\left[\mathcal{A}\left(\mathbf{x}_j\right)=\hat{\mathbf{x}}_j\right] \\
\leq & \exp \left(\varepsilon_1 \right) \cdot \operatorname{Pr}\left[\mathcal{A}\left(\hat{\mathcal{D}}_u \cap \hat{\mathcal{D}}_u^{\prime}\right)\subseteq \mathcal{O}\right] \cdot \operatorname{Pr}\left[\mathcal{A}\left(\mathbf{x}_j^{\prime}\right)=\hat{\mathbf{x}}_j\right] \\
=& \exp (\varepsilon_1) \cdot \operatorname{Pr}\left[\mathcal{A}\left(\hat{\mathcal{D}}_u^{\prime}\right) \subseteq \mathcal{O}\right],
\end{aligned}
\end{equation}
}
\end{proof}
\end{theorem}
where $\mathcal{O}$ denotes the subset of possible outputs. Eq.~\eqref{eq:theorem} indicates that $\mathcal{A}(\hat{\mathcal{D}}_u)$ satisfies $\varepsilon_1$-differential privacy according to the definition of differential privacy~\cite{dwork2014algorithmic}. Further, according to the composition theorem~\cite{dwork2014algorithmic}, our approach finally satisfies $(|\hat{\mathcal{D}}_s|\varepsilon_0^2+\varepsilon_0\sqrt{-2|\hat{\mathcal{D}}_s|\log{\delta}}+\varepsilon_1,\delta)$-differential privacy and gives the differential privacy guarantee.

\section{Experiments}
To verify the effectiveness of our proposed discriminative-generative distillation approach \textbf{DGD}, we conduct experiments on four datasets (MNIST~\cite{lecun2010mnist}, Fashion-MNIST~\cite{fashionMnist} (FMNIST), SVHN~\cite{2011SVHN} and CIFAR-10~\cite{cifar10}) and perform comparisons with 13 state-of-the-art benchmarks, including 6 explicit approaches that train models with generative data (DP-GAN~\cite{xie2018differentially}, PATE-GAN~\cite{jordon2018pate}, GS-WGAN~\cite{chen2020gs}, G-PATE~\cite{long2019scalable} and DP-MERF~\cite{harder2020differentially}, DataLens~\cite{wang2021ccs}), and 7 implicit approaches that train models with differentially private learning (DPSGD~\cite{abadi2016deep}, zCDP~\cite{yu2019differentially}, GEDDP~\cite{nasr2020improving}, DP-BLSGD~\cite{chen2020stochastic}, RDP~\cite{Yu2021icml}, TSADP~\cite{papernot2021tempered} and GEP~\cite{yu2021iclr}). Here, all explicit approaches but DP-GAN apply teacher-student learning to distill models, while all implicit approaches perform differentially private learning without model distillation. To make the comparisons fair, our experiments use the same experimental settings as these benchmarks and take the results from their original papers. Note that original PATE~\cite{papernot2016semi} conducted experiments with private data to simulate public data, thus we just compare to it in component analysis experiment.

MNIST and FMNIST are both 10-class datasets containing 60K training examples and 10K testing examples. The examples are $28\times 28$ grayscale handwriting digit images or fashion images. SVHN is a real-world $32\times32$ color digit image dataset that contains $73257$ training examples, $26032$ testing examples and $531131$ extra training examples. CIFAR10 consists of 60K $32\times32$ color images in 10 classes, including 50K for training and 10K for testing.

For each dataset, we take its training examples as private data and directly learn a baseline classifier as the discriminator as well as an ensemble of teachers, and then transfer the teacher knowledge to learn student. We set the Laplacian noise scale to be $2/\varepsilon_0=40$. We generate the same number of synthetic data as the private training data with the learned generator by randomly generating latent codes and then feeding into the generator to require synthetic data, \eg, generating 60K synthetic images for MNIST. In VAE reconstruction, we set $c=32$ and $\varepsilon_1=0.01$. The models are evaluated on testing examples with privacy cost, classification accuracy and accuracy drop with respect to baseline. 

We mainly use simple network structures that are the same to the benchmarks for teachers and student to conduct the experiments. On MNIST and FMNIST, the networks of baseline and teachers have the same structure, which contains two $3\times3$ convolutional layers (with ReLU activation and max-pooling, and 64 and 128 channels, respectively) to extract features hierarchically, followed by the softmax output layer indicating 10 classes. {Each convolutional layer has the stride of 1 with 1-padding and are randomly initialized with Xavier.} On SVHN, we add two extra fully-connected layers (with 384 and 192 neurons) with ReLU. {We use Adam optimization algorithm to learn all models, and set batch size as 128. To learn all teachers, the iteration rounds are 3000, the learning rate is first set to 0.05 and decays linearly with iteration round to 0. For generator learning, the iteration rounds are 200, the learning rate is first set to 0.2 and 10 times decays every 80 rounds. To learn VAE and student, the iteration rounds are 500, the learning rate is first set to 0.001 and decays linearly with iteration round to 0.} The teacher number on these three datasets is 250. We also study a complex structure for teachers and conduct an experiment on CIFAR10 with 100 teachers. Here, we fine-tune a vision transformer~\cite{dosovitskiy2020image} {on CIFAR10 training set} and modify the dimension of the last fully-connected layer to 10. {The model is pretrained on ImageNet and gives a top-1 classification accuracy of 81.4\%.}

\subsection{State-of-the-Art Comparison}
We conduct comparisons with 6 explicit approaches under different privacy budget on MNIST and FMNIST and  7 implicit approaches on CIFAR10. The performance is evaluated with test accuracy of student and accuracy drop with respect to its baseline under the condition of $(\varepsilon,\delta)$-differential privacy. Here, $\varepsilon$ is privacy budget and $\delta$ is failure probability. A lower privacy budget means a stronger privacy guarantee. 

\myPara{Comparisons with 6 Explicit Approaches.}~In the comparisons, we check the performance under different privacy budget, and report the results in Tab.~\ref{tab:mnist_and_fmnist}. Our approach takes 1300 queries under $\varepsilon=10.0$ and 27 queries under $\varepsilon=1.00$, respectively. All approaches are under a low failure probability $\delta=10^{-5}$. The test accuracy of baseline model is 99.2\% on MNIST and 91.0\% on FMNIST. 

From Tab.~\ref{tab:mnist_and_fmnist}, under the same condition of high privacy budget, we can see that our student achieves the highest test accuracy of 97.4\% on MNIST and 88.2\% on FMNIST, which remarkably reduces the accuracy drop by 1.80\% and 2.80\% respectively. It shows that our approach has the best privacy-preserving ability and minimal accuracy drop. 
Under the same low privacy budget, all approaches suffer from accuracy drop with respect to their counterparts under high privacy budget. However, our student still delivers the highest test accuracy and the lowest accuracy drop on both datasets. These results imply that discriminative stream plays an important role in knowledge transfer from private data. First, discriminative stream provides class identity supervision thus we cannot just use generative stream. Second, it uses certain queries to balance privacy protection and model accuracy. 
\begin{table}[!htbp]
 	\setlength{\tabcolsep}{10.5pt}%
 	\renewcommand\arraystretch{1.3}
	\begin{center}
		\begin{threeparttable}
		\caption{Test accuracy comparisons with 7 implicit approaches on CIFAR10 under different $\varepsilon$  ($\delta=10^{-5}$).}
			\begin{tabular}{c|cc}
				\hline
				Approach&$\varepsilon$&Accuracy~(\%)\cr
				\cline{1-3}
				DPSGD~\cite{abadi2016deep}&3.19&60.7\cr
				zCDP~\cite{yu2019differentially}&6.78&44.3\cr
				GEDDP~\cite{nasr2020improving}&3.00&55.0\cr
				DP-BLSGD~\cite{chen2020stochastic}&8.00&53.0\cr
				RGP~\cite{Yu2021icml}&8.00&63.4\cr
				TSADP~\cite{papernot2021tempered}&7.53&66.2\cr
				GEP~\cite{yu2021iclr}&5.00&70.1\cr
				\textbf{Our DGD}&{\bf 3.00}&{\bf 73.6}\cr
				\hline
			\end{tabular}
			\label{tab:cifar10}
		\end{threeparttable}
	\end{center}
\end{table}

\begin{figure*}[t]
	\centering
	\includegraphics[width=0.245\linewidth]{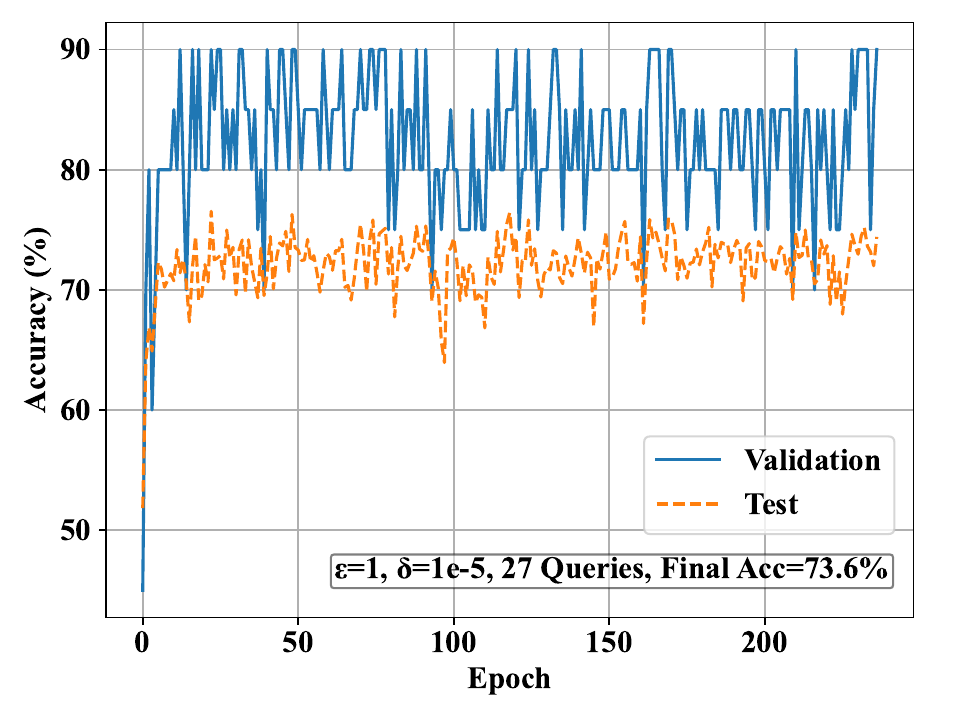}
	\includegraphics[width=0.245\linewidth]{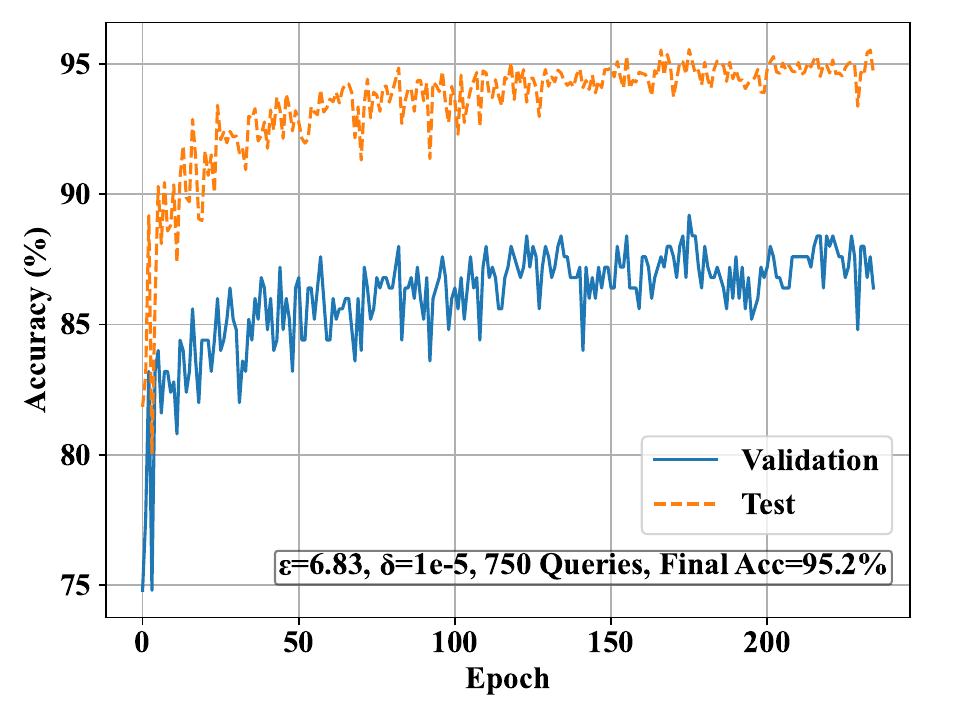}
	\includegraphics[width=0.245\linewidth]{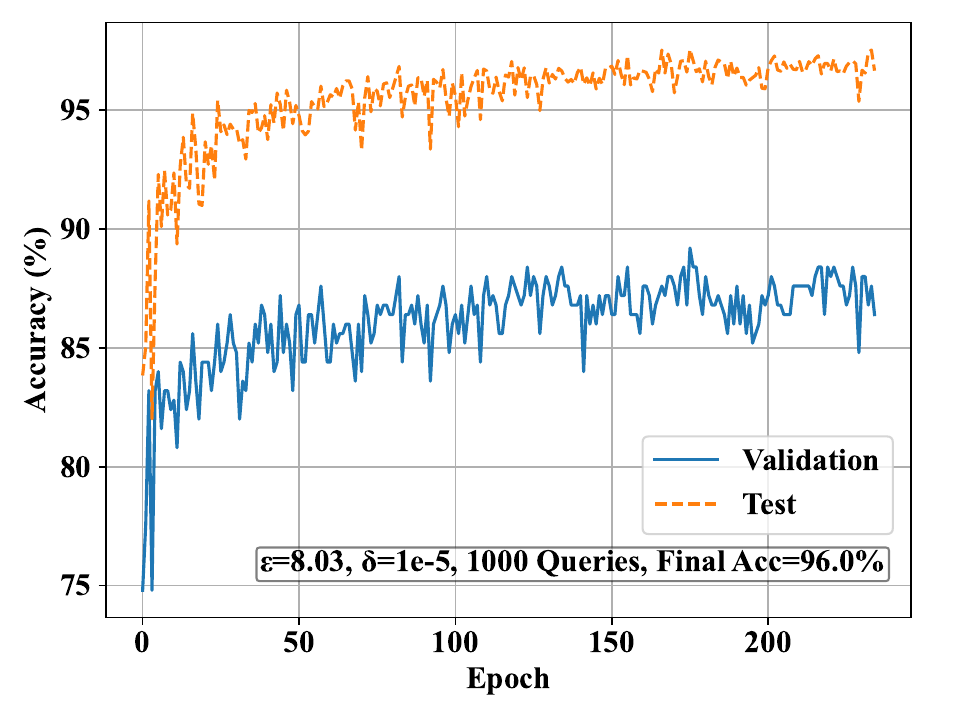}
	\includegraphics[width=0.246\linewidth]{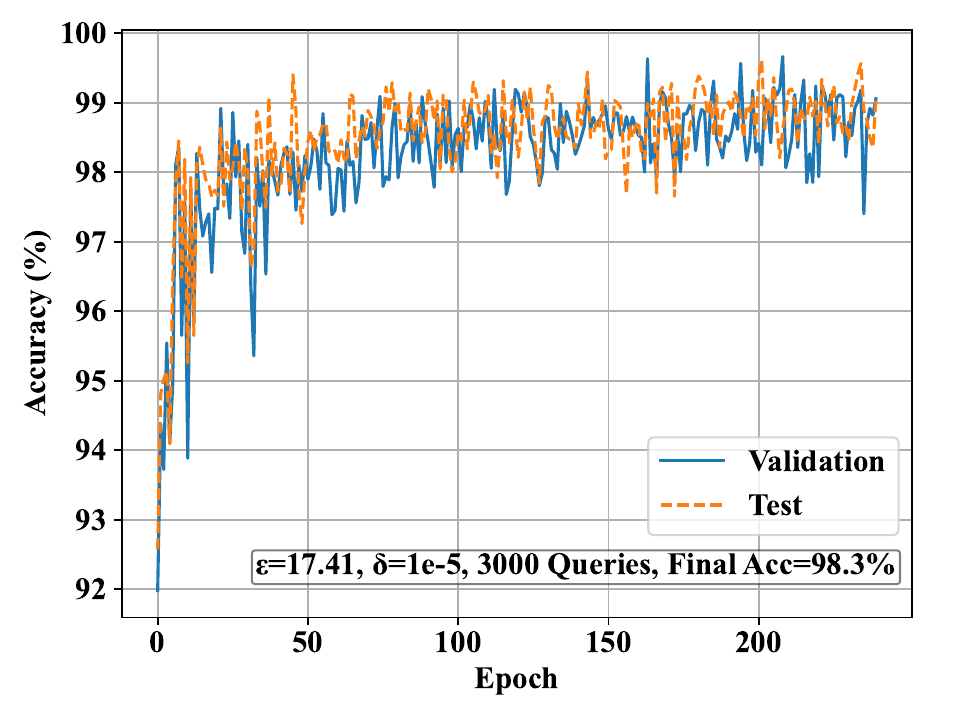}
	\caption{Model accuracy during training and privacy cost under different queries}
	\label{fig:train_loss}
\end{figure*}

\myPara{Comparisons with 7 Implicit Approaches.}~In addition, we conduct experimental comparison on CIFAR10 and report in Tab.~\ref{tab:cifar10}, where our approach achieves the highest accuracy of 73.6\% under the lowest privacy budget of 3.00. The main reason comes from that our approach adopts an extra generative stream to enhance knowledge transfer with massive synthetic data generated by a data-free learned generator. In this way, the missing knowledge can be recovered from generative stream and the accuracy can be improved.

\subsection{Component Analysis}
After the promising performance achieved, we further analyze the impact of each component in our approach, including label query, generator learning, teacher ensemble, noisy aggregation, and VAE-based regularization.

\myPara{Label Query.}~To study query effect on the trade-off between model accuracy and privacy protection, we compare the student learning under 27, 750, 1000 and 3000 queries. We treat the label of a generated example by the teacher ensemble as a query. The query number determines the privacy budget and failure probability, and we use differential privacy with moments accountant~\cite{papernot2016semi} as metric. More queries will cost a larger privacy budget and fixed query number will lead to constant privacy cost. The results are shown in Fig.~\ref{fig:train_loss}. They are as expected where a higher privacy budget leads to a higher model accuracy. Besides, in our approach, the private information that the delivered student can directly access is the noisy teachers' prediction outputs who pass through Laplacian aggregation. The results also reveal that our student learning by dicriminative-generative distillation can be  performed robustly and consistently under different label queries and providing a certain number of examples (\eg, 750) can lead to an impressive accuracy of 95.2\%. It is very helpful in many practical applications where a few of samples are available for sharing. Therefore, our approach can effectively learn privacy-preserving student models and control accuracy drop. 

\begin{figure}[t]
	\centering
	\includegraphics[width=1.0\linewidth]{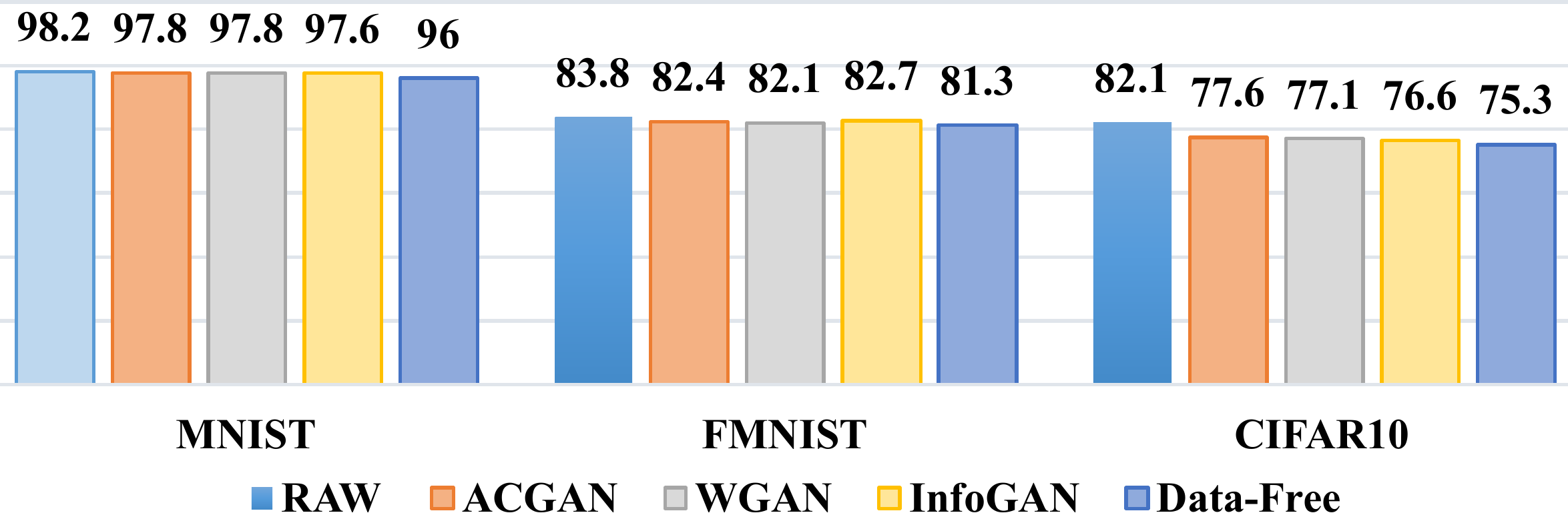}
	\caption{The effect of different generators on student accuracy (\%). Here, RAW means learning with private data.}
	\label{fig:effectgenerator}
\end{figure}

\begin{figure}[t]
	\centering
	\includegraphics[width=1.0\linewidth]{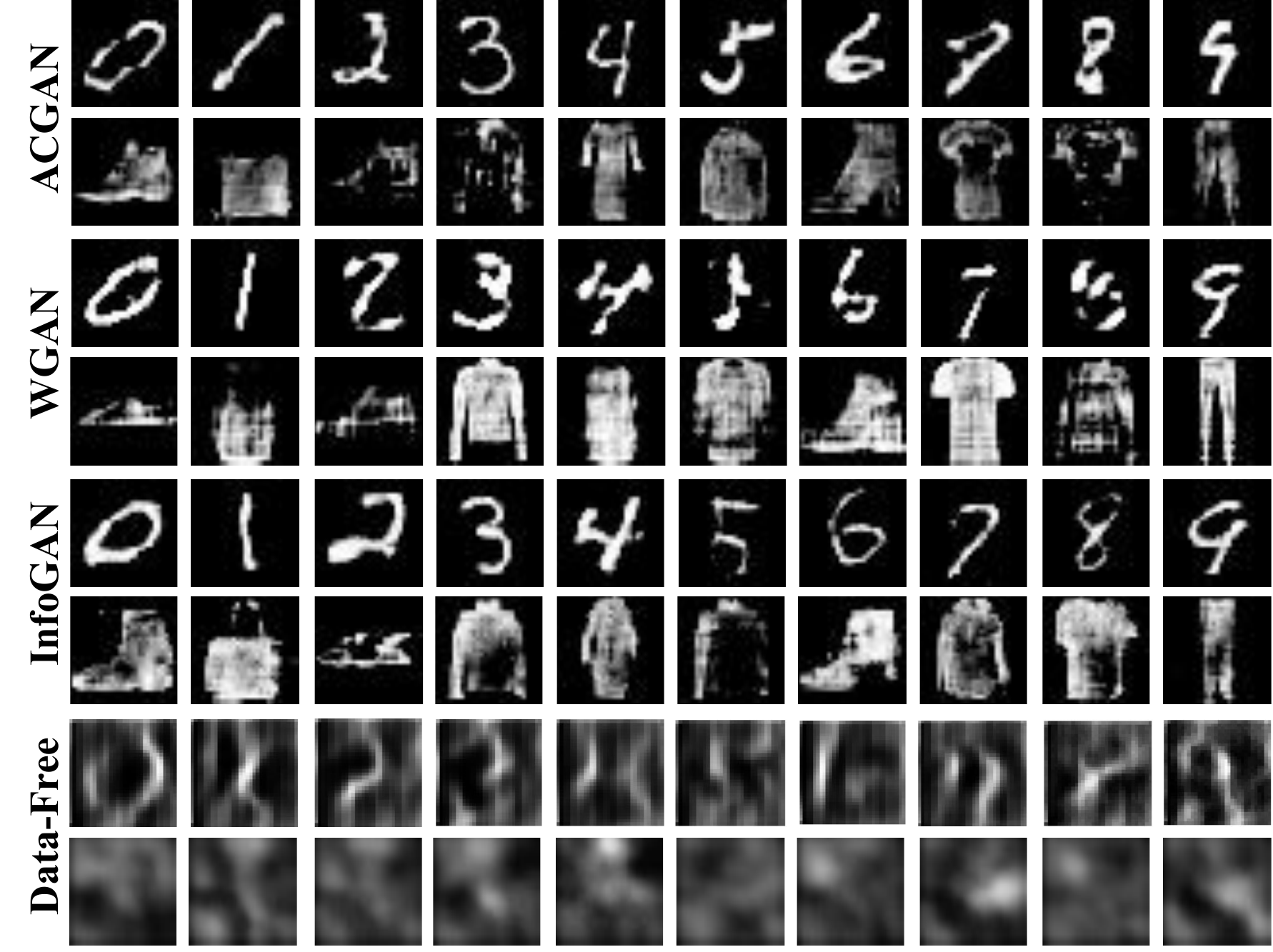}
	\caption{The generated images with different generator learning approaches. It is obvious that the examples generated by data-free learned generator protect privacy better.}
	\label{fig:generator}
\end{figure}

\myPara{Generator Learning.}~To study the effect of data generation, we conduct student learning on MNIST, FMNIST and CIFAR10 with the raw private data as well as synthetic data generated by four generative approaches, including ACGAN~\cite{odena2017conditional}, WGAN~\cite{arjovsky2017wasserstein}, InfoGAN~\cite{chen2016infogan} and our Data-Free learned generator. The results are shown in Fig.~\ref{fig:effectgenerator} and some generated examples can be seen in Fig.~\ref{fig:generator}. It is easy to distinguish the images generated by ACGAN, WGAN and InfoGAN, implying that these generators learned with private data may expose data privacy in spite of achieving higher accuracy. By contrast, the synthetic images generated by data-free learned generator are hardly identified by human. Thus, the data-free learned generator effectively protects privacy while delivering comparable accuracy since it matches the distribution of private data in discriminative space. 

\begin{figure}[t]
	\centering
	\includegraphics[width=1.0\linewidth]{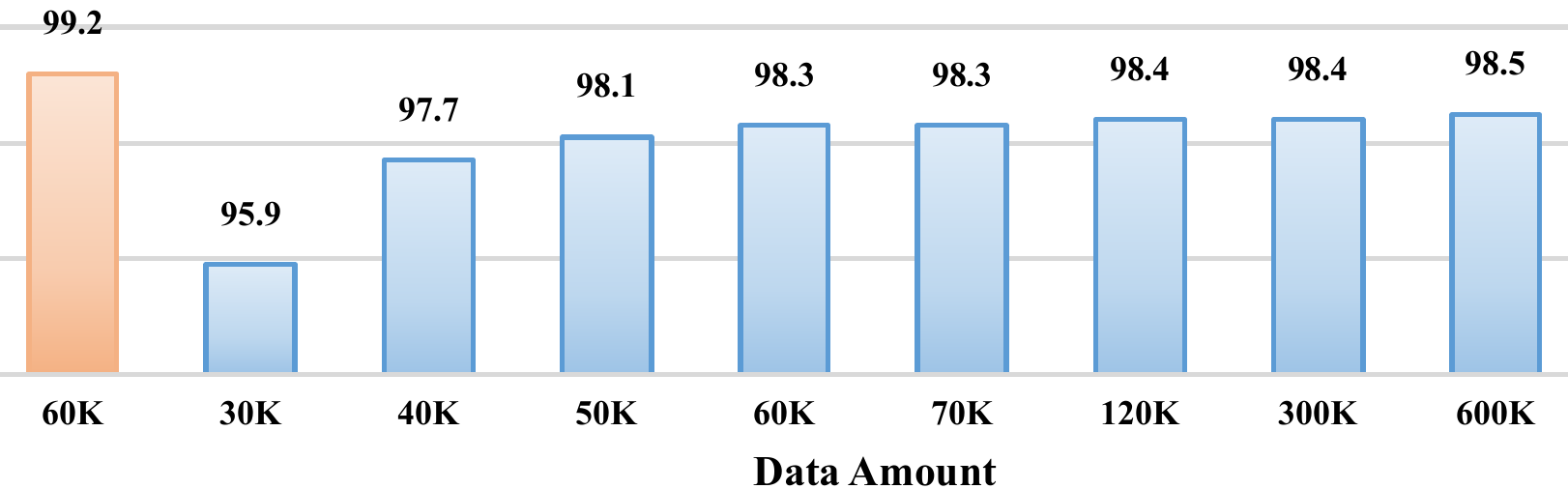}
	\caption{The accuracy~(\%) under different amounts of synthetic data, which  is generated with the generator trained by taking the baseline (in orange) as the fixed discriminator. The baseline is trained on 60K private training data of MNIST.}
	\label{fig:amount-acc}
\end{figure}

Beyond the generator learning method, we further check the impact of synthetic data. Towards this end, via the generator trained with the baseline on MNIST as the fixed discriminator, we generate 8 synthetic datasets with various amounts to train students and report their performance in Fig.~\ref{fig:amount-acc}. We can find that the model accuracy increases by training on more synthetic data and gets smooth after the used synthetic data reaches 60K that is equal to the number of private training examples. Therefore, we generate the same number of synthetic data as the private training data to provide a good trade-off between model performance and training efficiency.

\begin{figure}[t]
	\centering
	\includegraphics[width=0.49\linewidth]{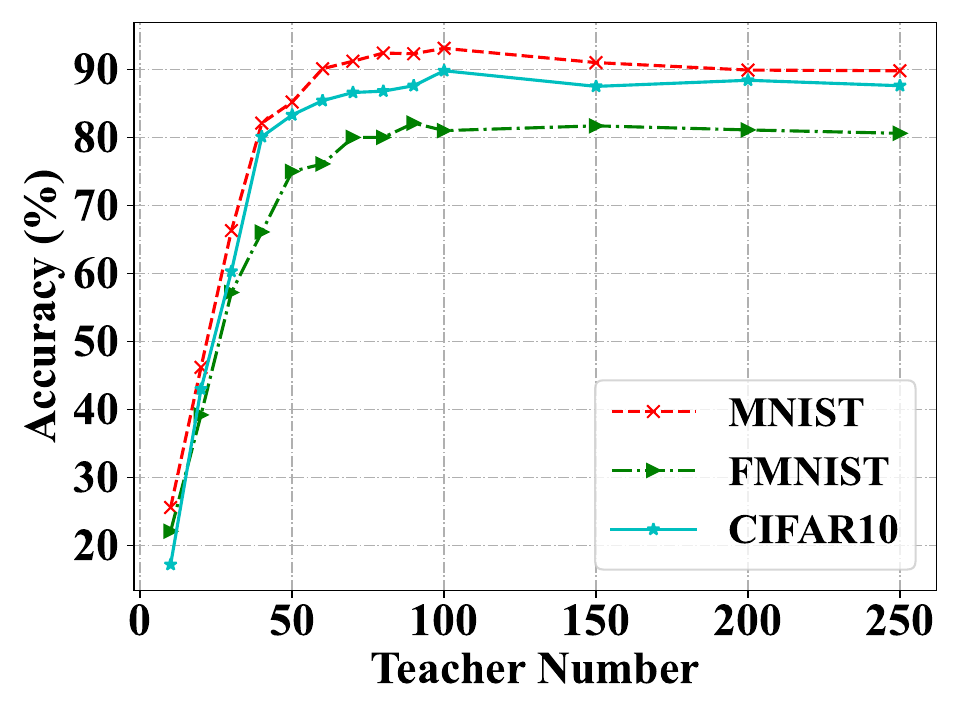}
	\includegraphics[width=0.49\linewidth]{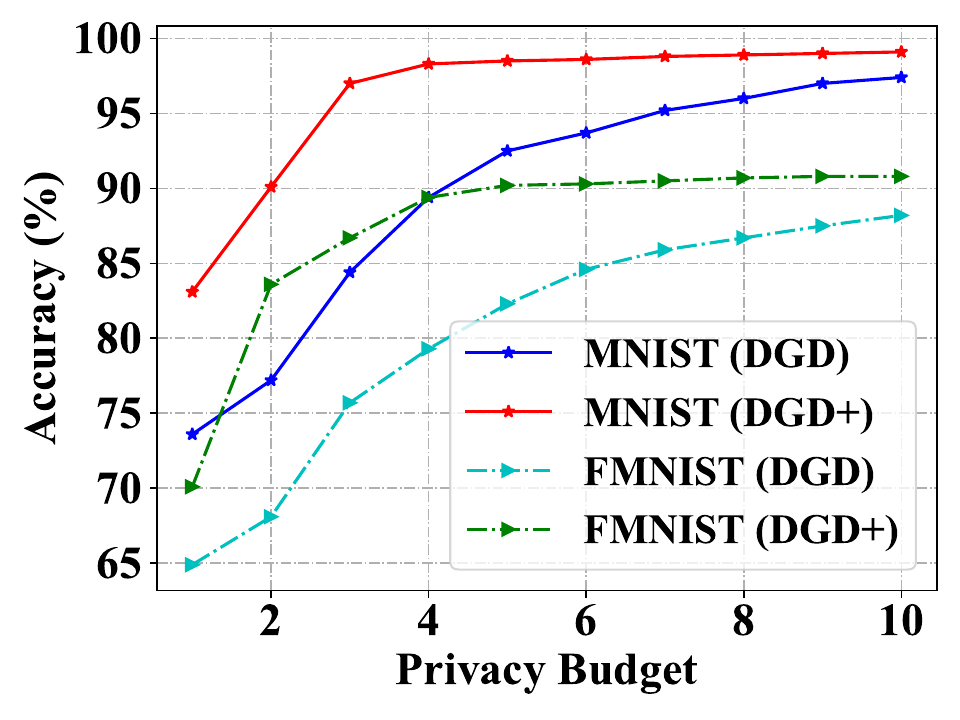}
	\includegraphics[width=0.49\linewidth]{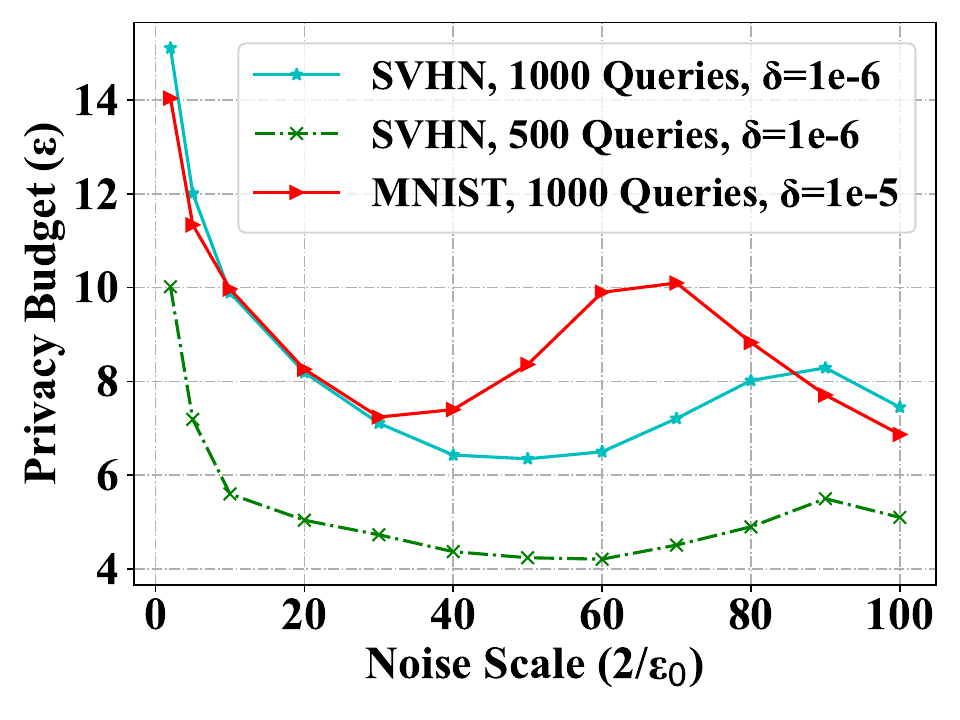}
	\includegraphics[width=0.49\linewidth]{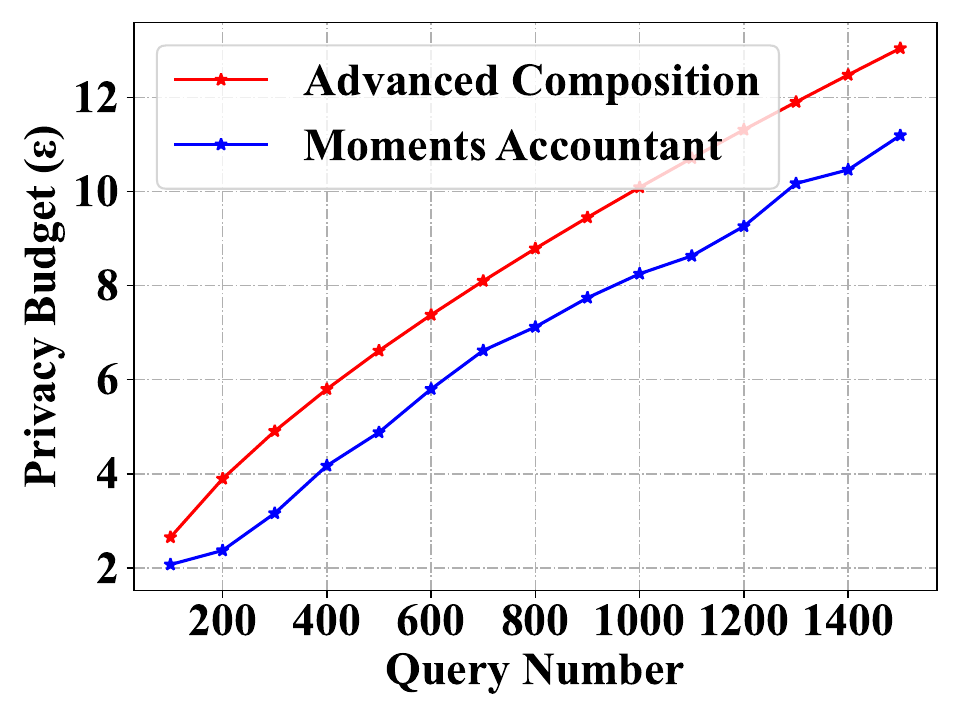}
	\caption{Top left: The privately aggregated accuracy of teacher ensemble under different teacher numbers. Top right: The student accuracy under different privacy budget $\varepsilon$ ($\delta=10^{-5}$). Bottom left: Privacy budget under different noise scale. Bottom right: Privacy budget under different query.}
	\label{fig:privacy_and_query}
\end{figure}

\myPara{Teacher Ensemble.}~We check the effect of teacher number on the accuracy of teacher ensemble. The top left of Fig.~\ref{fig:privacy_and_query} shows the results on three datasets for evaluating the effect on simple and complex classification tasks. We find that the accuracy increases along with teacher number within a certain range, indicating that the model performance can be boosted by increasing teacher number in a certain range. It is very helpful in real-world applications like federated learning~\cite{FL2019,swarm2021nature} where the private model can be improved by adding the sharing data parties. The performance starts to degrade when the teacher number increases to a certain value. Then the amount of partitioned training data for each teacher starts to become inadequate for learning, suggesting careful selection of the teacher number.

\begin{figure}[t]
	\centering
	\includegraphics[width=1\linewidth]{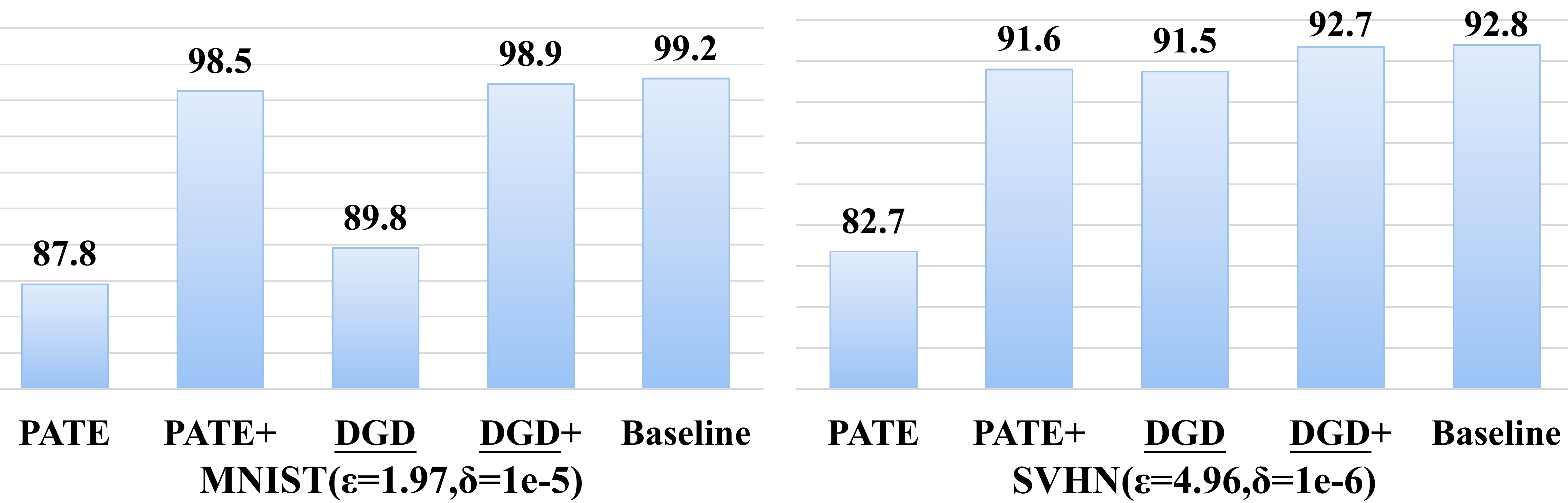}
	\caption{The effect of VAE on accuracy (\%) under different privacy budget and private aggregation mechanism.}
	\label{fig:VAE}
\end{figure}

\begin{figure*}[ht]
	\centering
	\includegraphics[width=1\linewidth]{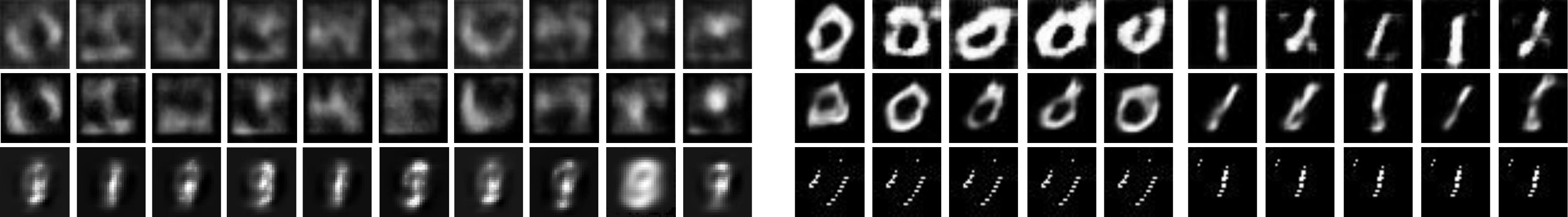}
	\caption{The left shows some reconstructed images with different reconstruction attacks~\cite{chen2019data} (Top),~\cite{fredrikson2015ccs} (Middle) and~\cite{yang2019ccs} (Bottom) on our student learned on MNIST. The right gives the results of model inversion attack~\cite{yang2019ccs} against binary classification models trained on the subset of MNIST only containing 0 and 1 with DataLens (Top), GS-WGAN (Middle) and DGD (Bottom).}
	\label{fig:attacks}
\end{figure*}

\myPara{VAE-based Regularization.}~To check the effect of VAE on student learning, we modify our approach for comparing to PATE~\cite{papernot2016semi} with Laplacian aggregation as well as its improved variant PATE+~\cite{papernot2018scalable} with Gaussian aggregation under the same experimental settings. Towards this end, we remove the generator and feed private training data (simulating public unlabeled data like~\cite{papernot2016semi,papernot2018scalable}) into VAE to learn students with Laplacian and Gaussian aggregation, leading to two modified approaches denoted as \underline{DGD} and \underline{DGD}+, respectively. We train various students on MNIST and SVHN under different privacy budget and conduct the comparisons. In our experiments, a few of training data serve as queries in noisy aggregation and the remaining most of the training data are fed into VAE where each example is reconstructed into a synthetic triple. The results are reported in Fig.~\ref{fig:VAE}, where using VAE in our \underline{DGD} and \underline{DGD}+ can consistently improve model accuracy over PATE and PATE+ without sacrificing privacy guarantee, respectively. For example, \underline{DGD}+ delivers an accuracy of 92.7\% on SVHN that is very close to 92.8\% achieved with baseline, implying the effectiveness of VAE-based regularization, since it can provide self-supervised knowledge enhancement to compensate the accuracy drop. We also achieve higher accuracy by Gaussian aggregation than Laplacian aggregation (\ie, PATE+ vs. PATE, and \underline{DGD}+ vs. \underline{DGD}) as stated in~\cite{papernot2018scalable}, implying the significance of noisy aggregation, introduced next.

\myPara{Noisy Aggregation.}~During the student learning, the obtained query labels are disturbed by noise with a scale of $2/\varepsilon_0$. Theoretically, the higher the noise scale is, the lower privacy cost and better privacy protection is. However, too high noise scale may cause label distortion, making student difficult to learn useful knowledge and unsuitable for practical deployment. We study how the noise scale $2/\varepsilon_0$ affects the privacy cost and report in the bottom left of Fig.~\ref{fig:privacy_and_query}. We can observe that the privacy cost declines rapidly with the noise scale within a certain range, has a short rise in the middle, and then keeps smooth. We suspect the main reason is the calculation process of privacy protection metric with moments accountant. Thus, we could select a noise scale (\eg, $2/\varepsilon_0=30$) to provide a good trade-off between privacy protection and model accuracy.

To further study the effect of noisy aggregation, we conduct experiments on MNIST and FMNIST and report the results in the top right of Fig.~\ref{fig:privacy_and_query} where DGD+ improves DGD with Gaussian aggregation. It shows that the improvement of all students' accuracy starts rapid and gets smooth when increasing the privacy budget. Moreover, as expected, the accuracy of students trained with Gaussian aggregation is remarkably improved over Laplacian aggregation, suggesting that more advanced noisy aggregation mechanisms can be incorporated into our framework to facilitate performance.

\subsection{Privacy-Preserving Analysis}
We further study the privacy-preserving ability of the learned students. Towards this end, we conduct both theoretical analysis and practical analysis. 

In theory, our approach trains students from VAE-reconstructed synthetic data in generative stream and noisy labels in discriminative stream, where the reconstructed synthetic data are achieved by inputting synthetic data into VAE and reconstructing from noisy latent codes. As discussed in \ref{sec:discussion}, it contains two parts of privacy budgets and totally achieves $(|\hat{\mathcal{D}}_s|\varepsilon_0^2+\varepsilon_0\sqrt{-2|\hat{\mathcal{D}}_s|\log{\delta}}+\varepsilon_1,\delta)-$differential privacy over $|\hat{\mathcal{D}}_s|$ queries for all $\delta \in (0,1)$. In our experiments, generative privacy budget $\varepsilon_1=0.01$ is very small and can be ignored in the total privacy budget. Discriminative privacy budget dominates the total privacy budget, \eg, having a much higher privacy budget of 5.80 under $\varepsilon_0=1/20$, $\delta=10^{-5}$ over 400 queries.
For discriminative stream, we first use differential privacy with advanced composition~\cite{dwork2014algorithmic} to track privacy loss, seeing the red curve in the bottom right of Fig.~\ref{fig:privacy_and_query}. To better track the privacy loss, we further use differential privacy with moment  accountant~\cite{abadi2016deep} and add an advanced limit~\cite{papernot2016semi} to get a lower privacy budget. The results can be seen the blue curve in the bottom right of Fig.~\ref{fig:privacy_and_query}.
We can see that the privacy guarantee is satisfied in all metrics, \eg, a privacy budget of 10.1 with advanced composition or 8.03 with moments accountant under 1000 queries, leading to an effective trade-off between privacy-preserving model learning and accuracy drop control. 

In practice, a learning process of our approach can achieve several major models, including: 1) the baseline model (serving as the fixed discriminator) and the ensemble of teachers that are kept privately, 2) the data-free learned generator that can be delivered to provide valuable synthetic data for training more further models in a more privacy-preserving manner than using private or other generative data (as shown in Fig.~\ref{fig:generator} and discussed above), and 3) the student that is delivered for privacy-preserving deployment. 
We take the student learned on MNIST under $(10.0,10^{-5})$-differential privacy as an example and study its privacy-preserving ability against three reconstruction attacks, including reconstruction with data-free learned generator~\cite{chen2019data}, model inversion attack with confidence information and basic countermeasures~\cite{fredrikson2015ccs}, and adversarial model inversion attack with background knowledge alignment~\cite{yang2019ccs}. Some reconstructed results by these three attacks are shown in the left of Fig.~\ref{fig:attacks}, from the first to the third row, respectively. We can see that the reconstructed images are very different from the original images and hardly distinguished which number they are by human. Thus, the student can well protect data privacy whilst delivering a high accuracy of 97.4\%. To further verify the privacy-preserving ability of our students, we investigate an especial case by conducting inversion attack~\cite{yang2019ccs} against binary classification models that are trained on a subset of MNIST containing only `0's and `1's with GS-WGAN, DataLens and our DGD. The results are shown in the right of Fig.~\ref{fig:attacks}. The reconstruction results of GS-WGAN and DataLens can be distinguished by human, while our model can provide better protection against reconstruction attacks. From these results, we can safely claim that our approach can provide effective privacy-preserving model learning and the learned models are suitable particularly for practical application on privacy-conscious scenarios. 

\section{Conclusion}
Deep models trained on private data may pose the risk of privacy leakage. To facilitate model deployment, we proposed a dicriminative-generative distillation approach to learn privacy-preserving student networks. The approach takes dicriminative and generative models as bridge to distill knowledge from private data and transfer it to learn students in a semi-supervised manner. The supervised learning from noisy aggregation of multiple teachers can provide privacy guarantee, while the unsupervised learning from massive synthetic generated by a data-free learned generator can reduce accuracy drop. Extensive experiments and analysis were conducted to show the effectiveness of our approach. In the future, we will devise more advanced differential privacy mechanisms to improve the approach and explore the approach in more real-world applications like federated learning on medical images.  

\myPara{Acknowledgements.}~This work was partially supported by grants from the Beijing Natural Science Foundation (19L2040), National Key Research and Development Plan (2020AAA0140001), and National Natural Science Foundation of China (61772513). 
\bibliographystyle{IEEEtran}
\bibliography{bibieee}

\begin{IEEEbiography}[{\includegraphics[width=1in,height=1.25in,clip,keepaspectratio]{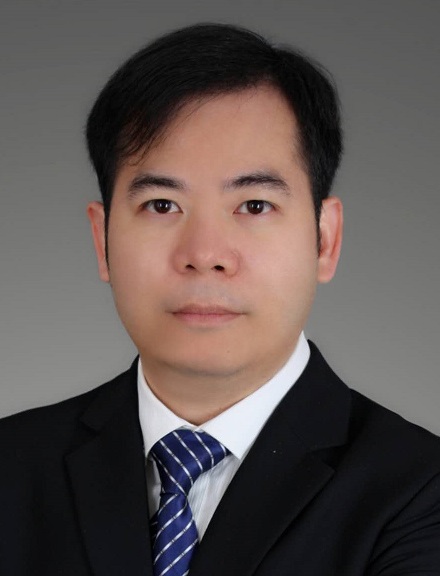}}]{Shiming Ge} (M'13-SM'15) is a professor with the Institute of Information Engineering, Chinese Academy of Sciences. Prior to that, he was a senior researcher and project manager in Shanda Innovations, a researcher in Samsung Electronics and Nokia Research Center. He received the B.S. and Ph.D degrees both in Electronic Engineering from the University of Science and Technology of China (USTC) in 2003 and 2008, respectively. His research mainly focuses on computer vision, data analysis, machine learning and AI security, especially trustworthy learning solutions towards scalable applications. He is a senior member of IEEE, CSIG and CCF.
\end{IEEEbiography}

\begin{IEEEbiography}[{\includegraphics[width=1in,height=1.25in,clip,keepaspectratio]{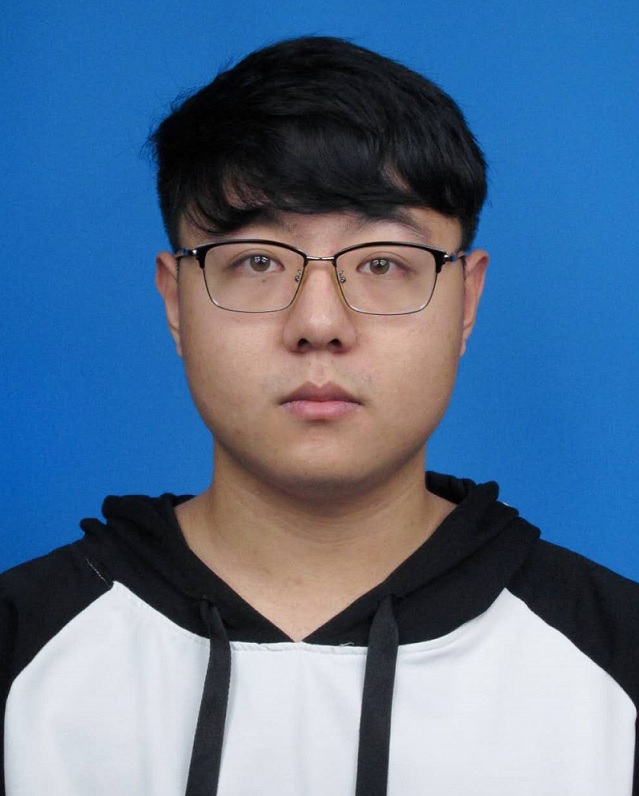}}]{Bochao Liu} received his B.S. degree in Electronical Information Science and Technology from the School of Information Science and Engineering in Shandong University, China. He is now a Ph.D Candidate at the Institute of Information Engineering at Chinese Academy of Sciences and the School of Cyber Security at the University of Chinese Academy of Sciences, Beijing. His major research interests are private-privacy machine learning.
\end{IEEEbiography}

\begin{IEEEbiography}[{\includegraphics[width=1in,height=1.25in,clip,keepaspectratio]{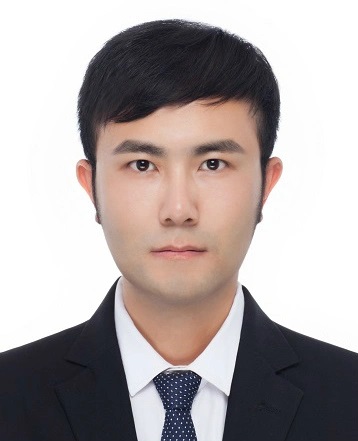}}]{Pengju Wang} is an Assistant Professor with the Institute of Information Engineering, Chinese Academy of Sciences. He received the B.S. degree from the School of Information Science and Engineering at Shandong University and M.S. degree from the School of Electronic Engineering at Beijing University of Posts and Telecommunications. His research interests include AI security and federated learning.
\end{IEEEbiography}

\begin{IEEEbiography}[{\includegraphics[width=1in,height=1.25in,clip,keepaspectratio]{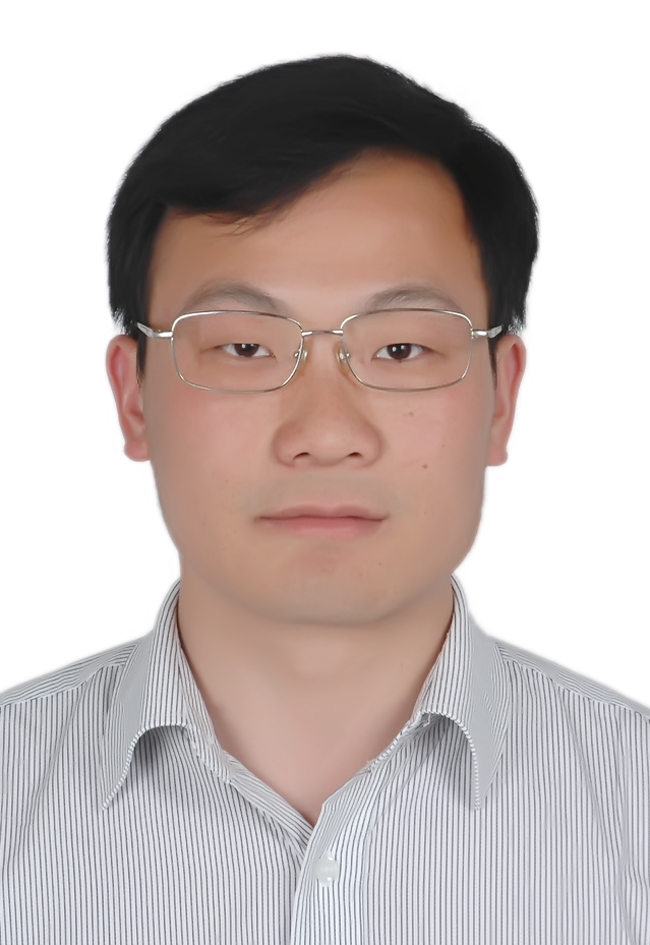}}]{Yong Li} is an Associate Professor with the Institute of Information Engineering, Chinese Academy of Sciences. He received the B.S. degree from the School of Computer Sciences at the Beijing Jiaotong University and Ph.D degree from the Institute of Computing Technology, Chinese Academy of Sciences. His research interests include security data analysis and the design of private machine learning methods and systems.
\end{IEEEbiography}

\begin{IEEEbiography}[{\includegraphics[width=1in,height=1.25in,clip,keepaspectratio]{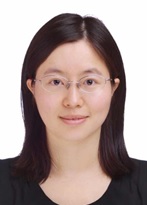}}]{Dan Zeng} (SM'21) received her Ph.D. degree in circuits and systems, and her B.S. degree in electronic science and technology, both from University of Science and Technology of China, Hefei. She is a full professor and the Dean of the Department of Communication Engineering at Shanghai University, directing the Computer Vision and Pattern Recognition Lab. Her main research interests include computer vision, multimedia analysis, and machine learning. She is serving as the Associate Editor of the IEEE Transactions on Multimedia and the IEEE Transactions on Circuits and Systems for Video Technology, the TC Member of IEEE MSA and Associate TC member of IEEE MMSP.
\end{IEEEbiography}

\vfill

\end{document}